\documentclass[10pt]{article}

\usepackage{hyperref}
\usepackage{url}
\usepackage{graphicx}
\usepackage{authblk}

\begin{document}

\title{OrigamiSet1.0: Two New Datasets for Origami Classification and Difficulty Estimation}

\author[1]{Daniel Ma}
\author[1,2]{Gerald Friedland}
\author[1,2]{Mario Michael Krell}
\affil[1]{University of California, Berkeley}
\affil[2]{International Computer Science Institute, Berkeley}

\maketitle

\begin{abstract}
Origami is becoming more and more relevant to research.
However, there is no public dataset yet available and there hasn't been
any research on this topic in machine learning.
We constructed an origami dataset using images 
from the multimedia commons and other databases.
It consists of two subsets: one for 
classification of origami images 
and the other for difficulty estimation.
We obtained $16000$ images for classification 
(half origami, half other objects) 
and $1509$ for difficulty estimation with $3$ different categories 
(easy: $764$, intermediate: $427$, complex: $318$).
The data can be downloaded at:
\url{https://github.com/multimedia-berkeley/OriSet}.
Finally, we provide machine learning baselines.
\end{abstract}

\section{Introduction}

\textit{Origami} is the art of paperfolding. 
The term arises specifically from the Japanese tradition, 
but that tradition has been practiced around the world for
several centuries now~\cite{Hatori2011}. In addition to being a recreational
art form, in recent years, origami has increasingly often been
incorporated into the work and
research of mathematicians and engineers. 

In the 1970s for example to solve the problem of packing large, flat
membrane structures to be sent into space, Koryo Miura used origami: he
developed a method for collapsing a large flat sheet to a much smaller area so
that collapsing or expanding the sheet would only require pushing or pulling at
the opposite corners of the sheet \cite{Miura1994}.
Similarly, Robert Lang
is using computational origami-based research to help the 
Lawrence Livermore National Laboratory to
design a space telescope lens (``Eyeglass'') that can collapse 
down from $100$ meters in diameter to a size that will fit 
in a rocket with roughly
$4$ meters in diameter~\cite{Hyde2002,lang2008flapping}. 
In the field of mathematics, Thomas Hull is investigating enumeration 
of the valid ways 
to fold along a crease pattern (i.e., a diagram containing all the creases 
needed to create a model) 
such that it will lie flat. 
He uses approaches from coloring
in graph theory to solve the problem~\cite{Hull2015}.
In medicine, Kuribayashi et al.\ used
origami to design a metallic heart stent that can be easily threaded
through an artery before expanding where 
it needs to be deployed~\cite{Kuribayashi2006}.

In other words, origami is becoming an established part of science.
To support research on origami,
we decided to generate a new large origami dataset.
Part of it comes from filtering it out from the multimedia commons
\cite{Bernd2015}.
The multimedia commons comprise information 
about a dataset of $99$ Million images 
and $800000$ videos that have been uploaded on 
Flickr under creative commons license (YFCC100M)~\cite{Thomee2016}.
This work is part of a bigger project that creates a
framework around this data to support multimedia big data field studies
\cite{mmbds}.

Datasets are crucial for new developments and major progress in machine learning. 
In particular, datasets of images have allowed researchers to make 
significant advances in the field of computer vision.  
For example, ImageNet 
~\cite{deng2009imagenet}, a 
dataset of millions of images and corresponding noun labels, has been a useful 
resource in creating and benchmarking large-scale algorithms for image 
classification.  The German Traffic Sign Detection Benchmark Dataset 
~\cite{houben2013detection} has a
practical use for self-driving vehicles to detect traffic signs in order for 
them to act appropriately.  The MNIST database ~\cite{lecun2010mnist}, a vast 
database of handwritten 
numeric digits, has been used for training and testing various classification 
techniques on image recognition. 

Here, we introduce two new datasets which we collectively call OrigamiSet1.0.  
The two datasets together consist of more than $15,000$ images.  In this paper, 
we first describe the properties of the datasets themselves 
in Section~\ref{s:oset}.
Next, we provide
baseline evaluations for distinguishing images of origami 
from images that contain no origami (Section~\ref{s:oc}) 
as well as image-based
difficulty estimation
of origami models (Section~\ref{s:ode}).
We conclude in Section~\ref{s:conc}.

\section{OrigamiSet1.0}
\label{s:oset}

First, we describe the dataset to distinguish origami images from normal images
and then we introduce the dataset for difficulty estimation.
The data can be downloaded at: 
\url{https://github.com/multimedia-berkeley/OriSet}.

\subsection{Origami Image Classification}

This dataset consists of $8,011$ images containing origami and $7,976$ images 
that do not contain origami. The majority of the origami images were 
scraped from two major origami databases: Oriwiki ~\cite{oriwiki} and 
GiladOrigami ~\cite{giladorigami}. 
Before scraping both websites for images, we 
contacted the administrators of both databases, asking for permission to scrape 
the images. The Oriwiki administrators are unreachable, but Gilad 
Aharoni gave us permission to use his images in our dataset (Gallery only). 

\subsubsection{Scraping procedure}
To scrape the 
images from Oriwiki, we found that each model's page was assigned a Model ID 
number included in the URL of the page, so we wrote a Python script that 
iterated through every Model ID number between $0$ and $99,999$, retrieving the 
image from the corresponding model's page.  Afterwards, we noticed that a 
significant portion of the retrieved images were placeholder images. So we 
removed those, which resulted in $3,934$ images from Oriwiki. 
As for GiladOrigami, 
unlike Oriwiki, each model image did not have its own page or model ID number.  
Instead, we went through the gallery section of the site and scraped all the 
origami images from each page in the gallery by hand. 
We note that the gallery does 
not contain all of the images of the site, but scraping from 
GiladOrigami has produced $2,140$ images of origami.

The remainder of 
the origami images were taken from the YFCC100M data browser~\cite{Kalkowski2015}. 
Starting with the
search term \textit{origami} resulted in over $13,000$ images, but more than 
$30\%$ of the images did not contain origami. 
Due to a tag cloud visualization,
we used 
\textit{papiroflexia} and \textit{origamiforum} as search terms which 
generated a more reliable, albeit 
smaller set of results.  After some minor hand-cleaning, which involved 
removing non-origami images as well as images containing people, we obtained 
$1,937$ origami images.

\subsubsection{Data properties and non-origami class creation}

\begin{figure}[ht]
	\centering
	\includegraphics[scale=0.3]{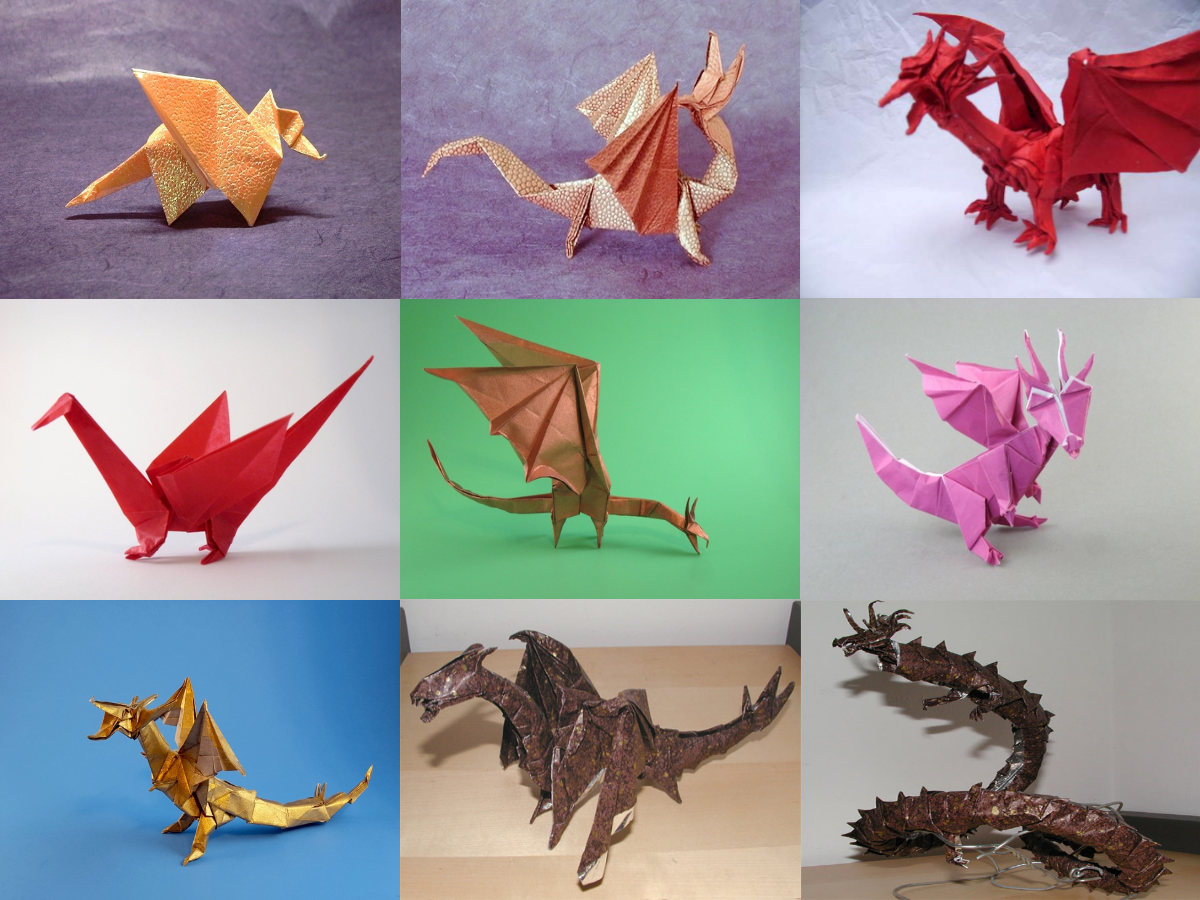}
	\caption{For some model topics, (such as dragons, Santa Claus, pandas, 
	etc.), there are multiple models that were meant to represent the same 
	topic.  This example contains $9$ different dragon models~\cite{giladorigami,oriwiki}.}
  \label{f:topics}
\end{figure}

In this section, the properties of the origami images in our dataset are 
described.  In our diverse set, each of the models itself is 
the main focus of the 
image. For example, all of these images do not contain humans or human body 
parts. Furthermore, there is generally only one model in the center of the image. In 
addition, there were many images of models that were meant 
to represent the same topic. For example, there were multiple instances of 
dragons; some were very simplistic, while others incorporated much more details 
in the wings, head, and appendages of the model (see Figure~\ref{f:topics}).

To explore the diversity of the dataset as well as its relation to the ImageNet 
dataset, we applied a VGG16 
neural network ~\cite{Simonyan2014} to the YFCC100M origami images
that was trained on ImageNet with $1000$ 
common classes.  ImageNet does not include the class \textit{origami}.

\begin{table}[ht]
\renewcommand{\arraystretch}{0.9}
\tabcolsep=0.11cm
\small
\begin{center}
\begin{tabular}{|l|r||l|r|c|c|c|c|c||c|c|}
\hline
\multicolumn{2}{|c||}{Top-1} &\multicolumn{2}{c|}{Top-5}\\ 
\multicolumn{1}{|l|}{name}&\multicolumn{1}{c||}{n}
&\multicolumn{1}{c|}{name} &\multicolumn{1}{c|}{n} 
\\ 
\hline 
pinwheel & 243 & envelope & 788\\
envelope & 243 & pinwheel & 518\\
carton & 117 & carton & 499\\
paper towel & 76 & packet & 328\\
honeycomb & 71 & handkerchief & 314\\
lampshade & 41 & paper towel & 302\\
rubber eraser & 39 & rubber eraser & 238\\
handkerchief & 37 & candle & 218\\
pencil sharpener & 35 & lampshade & 192\\
shower cap & 34 & wall clock & 168\\
\hline
\end{tabular}
\end{center}
\caption{Top-i predictions for our extracted YFCC100M subset ($n = $ number of 
occurrences).}
\label{t:top15}
\end{table}

We found that the top-1 predictions were spread among $263$ different classes, 
while the top-5 predictions were spread among $529$ classes. The most common 
classes that appeared are shown in Table \ref{t:top15}.  As one can tell, our 
origami images were very often classified as items that are similar to 
paper-folding and origami (such as \textit{envelope} or \textit{handkerchief}) 
as well as things that involve paper (such as \textit{paper towel} or 
\textit{carton}).  In some cases, some images were classified as objects that 
an origami model was meant to represent such as flowers, candles, or bugs.

With these labels in mind, we generated a non-origami class by using the 
ILSVRC2011 
validation data ~\cite{Russakovsky2014}. 
For each label, excluding 
\textit{pinwheel, 
envelope, carton, paper towel, packet,} and \textit{handkerchief}, we gathered 
the first 8 examples, producing $7976$ non-origami images.

\subsection{Difficulty Estimation}
\label{s:diff}

To the best of our knowledge, 
there is no single useful standard 
for defining the difficulty
of an origami model. 
Some heuristics to estimate the difficulty of folding a
model include the number of creases involved or the number of steps included in
the diagrams in the model. 
However, such data is not easily available for a large number of 
models and collecting such data might be infeasible. Instead, we estimate
difficulty based on the appearance of the model.

\begin{figure}[ht]
  \centering
  \includegraphics[scale=0.5]{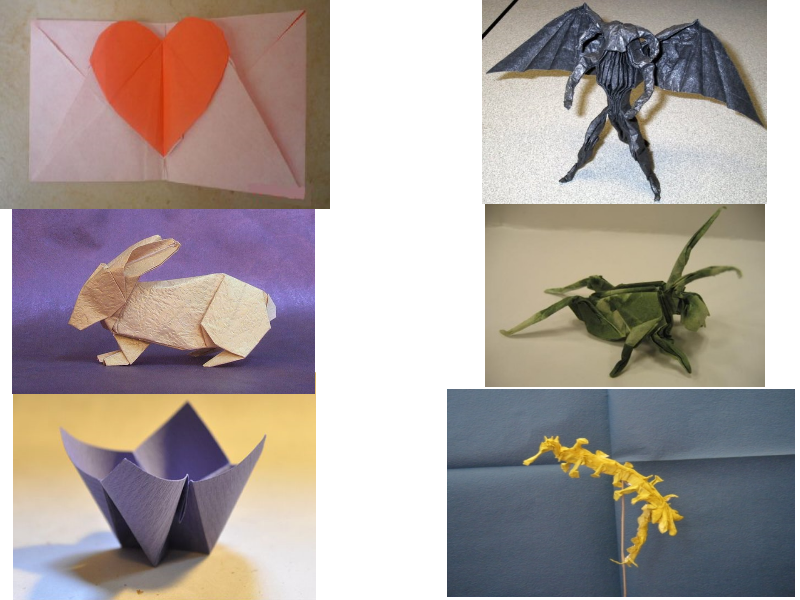}
	\caption{Examples of wrong difficulty assignment 
    in the Oriwiki database~\cite{oriwiki}.
    On the left are images of models that are ranked as very 
		difficult when they appear to actually be easy or intermediate models.  
		On 
		the right are images of models that are rated as very easy when 
		they 
		actually appear to be complex.}
	\label{fig:incorrect}
\end{figure}

At first, we used difficulty levels that were assigned to the origami images that we 
scraped from the Oriwiki database~\cite{oriwiki}. These difficulty levels were 
assigned by the uploading user.
Each difficulty label was a value 
between $1$ and $5$, where $1$ represents low difficulty and $5$ represents high 
difficulty. Again, we used the VGG16 features.
However this time, we trained an algorithm to predict the 
difficulty of an origami model. 
However, training a classifier could not 
produce any useful result (see Section~\ref{s:ode}). 
A likely reason for the low performance is that the labeling 
of difficulties by users is inconsistent or even wrong. 
This can be seen in Figure~\ref{fig:incorrect} 
where the images show a clear discrepancy between the difficulty 
label and the perceived difficulty.
Thus, we provide our own difficulty labels.

\begin{figure}[ht]
	\centering
	\includegraphics[width=0.95\textwidth]{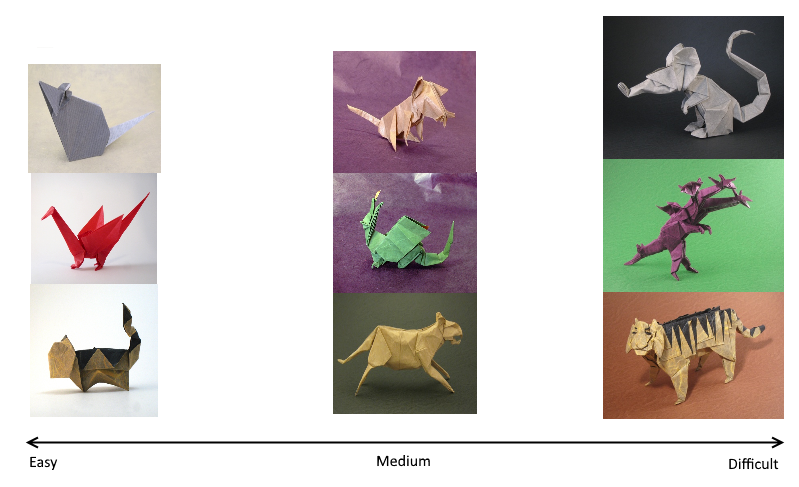}
	\caption{Examples of images from the GiladOrigami database~\cite{giladorigami}
	classified as easy, medium, and difficult (see also Table~\ref{t:diff})
  from three different topics: tiger, dragon, and rat.}
  \label{f:diff}
\end{figure}

We picked a subset of $1509$ images of origami from the set of images used for
origami image classification, hand-assigning each image a difficulty of 0
(easy), 1 (intermediate), or 2 (complex). In the end, there were $764$ easy
images, $427$ intermediate images and $318$ complex images. 
Several assumptions and
guidelines were followed in labeling the difficulty of the models in these
images. In particular, each image was assigned a difficulty based on the
assumption that the model in the image was made from a single sheet of paper.
Images containing modular models (i.e., models made from multiple pieces of paper) and 
tessellations (i.e., models with repeating patterns), as complex as they appear, usually 
involve folding an easy sequence of folds repetitively. Table~\ref{t:diff} contains 
general guidelines that were used for assigning a difficulty to each model
and Figure~\ref{f:diff} provides some examples.

\begin{table}[ht]
  \small
\begin{tabular}{|l|l|}
\hline
Easy (0) & - Contains very little detail\\
& - Appears very ``flat'', i.e. has very few layers of paper\\
& - Appearance may consist of a collection of simple geometric shapes\\
& - Has very few appendages, most of which are wide\\
& - Boxes and airplanes in most cases fall under this difficulty\\
\hline
Intermediate (1) & - Moderate amount of detail\\
& - Moderate number of appendages that are somewhat narrow\\
& - Inclusion of toes or fingers to appendages\\
& - Contain two objects in a single model, although one object\\ & may be significantly smaller than the other (e.g. man with violin)\\
& - Some shaping to add curves and a 3D appearance to the model\\
\hline
Complex (2) & - Highly detailed\\
&- Large number of appendages\\
&- High resemblance to the subject the model is supposed to represent\\
&- Contains several objects in a single model\\
&- If a model contains only two objects, the two objects are about \\
&the same size\\
&- Significant amount of shaping\\
&- Insects tend to fall in this category, especially if the models have \\
&long, narrow appendages\\
\hline
\end{tabular}
\caption{Guidelines for assigning difficulty levels (see also Figure~\ref{f:diff}).}
\label{t:diff}
\end{table}

One concern regarding our difficulty estimation is that in the process of 
hand-labeling difficulty, some models are actually more difficult or easier 
than they appear.  For example, a model with many appendages will be classified 
as difficult according to our heuristic, but the actual difficulty of the model 
may be intermediate instead.
Here, future contributions to the dataset are welcome to provide
corrections or new variants.
Note, however, that visual judgement of the difficulty cannot be as accurate
as a real estimated related to the folding procedure.

\section{Machine Learning Evaluation}
\label{s:ml}

Apart from the creation of the datasets, we also want to present and explore
two machine learning challenges on this data 
to encourage data analysis and algorithm development
for this kind of data.
It is meant to be seen as a first baseline and to quantify the challenges.

\subsection{Origami Classification}
\label{s:oc}

In the first setting, we want to distinguish origami from non-origami images.

\subsubsection{General Processing Setup}
For generating features of 
all our images we used the VGG16 neural network model implemented in MXNet.
In particular, each image's features were generated by taking the output
right before the last layer of the model. In addition, a list of labels was 
created to denote whether the resulting feature vector belongs to an origami image 
(1) or to a non-origami image (-1).  With these features and labels, we evaluated 
multiple classifier models using Scikit-learn's 
~\cite{scikit-learn} 5-fold 
cross-validation function.

\subsubsection{Classifier Comparison Results}
We compared multiple kernels of SVC models as well as the logistic regression 
model provided by scikit-learn \cite{scikit-learn}, all on 
default settings. Table~\ref{t:scores} shows 
the accuracy scores for each model.

\begin{table}[ht]
\centering
\caption{Origami classification estimation: 5-fold cross-validation scores for 
different models on default settings}
\label{t:scores}
\begin{tabular}{|l|l|}
	\hline
	Classifier & Accuracy Score\\ \hline
	SVC: Linear Kernel & $0.9674\pm 0.0049$\\ \hline
	SVC: RBF Kernel & $0.9705\pm 0.0051$\\ \hline
	SVC: Poly Kernel & $0.9746\pm 0.0063$\\\hline
	SVC: Sigmoid Kernel & $0.9215\pm 0.0128$\\\hline
	Logistic Regression & $0.9707\pm0.0071$\\\hline
\end{tabular}
\end{table}

\subsubsection{Discussion}

We note that a majority of the classifiers' accuracy scores are actually quite 
close, so it is difficult to say which classifier has the 
best performance.
However, the results indicate that a good classification is possible with our data
and that with further tuning, scores around $99\%$ should be possible.

\subsection{Origami Difficulty Estimation}
\label{s:ode}

Similar to the Origami Image Classification setup, we again used the output 
vector right before the last layer of the VGG-16 neural network as the feature 
for each image.  As before, we looked at how well several models on 
default settings could predict the
difficulty based on the values provided by the users of the Oriwiki \cite{oriwiki} 
database.  We performed a 5-fold cross-validation on the same classifiers 
that we used earlier in Section~\ref{s:oc}, but produced $R^2$ scores as well as 
balanced accuracy scores to 
account for the fact that the images are not distributed very evenly across the 
different difficulty labels~\cite{Straube2014}.

\begin{table}[ht]
	\centering
	\caption{Difficulty Estimation: 5-fold CV Scores of 
	Different Models using Oriwiki Labels}
	\label{t:owDiff}
	\begin{tabular}{|l|l|l|}
		\hline
		Classifier & Balanced Accuracy Score & $R^2$ Score\\ \hline
		SVC: Linear Kernel & $0.2775\pm0.0220$ & $-0.5751\pm0.0842$\\ \hline
		SVC: RBF Kernel & $0.4488\pm0.0451$ & $-0.5383\pm 0.1282$\\ \hline
		SVC: Poly Kernel & $0.3202\pm0.0170$ & $-0.4513\pm0.1022$\\ \hline
		SVC: Sigmoid Kernel & $0.2931\pm0.0470$ & $-0.737\pm0.0832$\\ \hline
		Logistic Regression & $0.3034\pm0.0233$ & $-0.4958\pm0.0879$ \\\hline
	\end{tabular}
\end{table}

Table~\ref{t:owDiff} clearly shows that useful difficulty estimation 
is not possible with the labels from the Oriwiki database.
For this reason, we decided to label a subset of the origami 
images by hand (see Section~\ref{s:diff}).  With these labels, we again fit the 
data to several models on default settings and got significantly better 
accuracy and $R^2$ scores (see Table~\ref{t:hlDiff}).

\begin{table}[ht]
	\centering
	\caption{Difficulty Estimation: 5-fold CV Scores of Different Models using 
	our Labels}
	\label{t:hlDiff}
	\begin{tabular}{|l|l|l|}
		\hline
		Classifier & Balanced Accuracy Score & $R^2$ Score\\ \hline
		SVC: Linear Kernel & $0.7581 \pm0.0151$ & $0.5631\pm0.0716$\\ \hline
		SVC: RBF Kernel & $0.7826\pm0.0217$ & $0.6293\pm0.1323$\\ \hline
		SVC: Poly Kernel & $0.7854\pm0.0172$ & $0.6158\pm0.0948$\\ \hline
		SVC: Sigmoid Kernel & $0.7223\pm0.0281$ & $0.5822\pm0.1516$\\ \hline
		Logistic Regression & $0.7625\pm0.0101$ & $0.5738\pm0.0789$\\ \hline
	\end{tabular}
\end{table}

\subsubsection{Evaluation setup}
To further improve on this classifier, we decided to tune the kernel and C 
hyperparameters of the SVC. Hence, we used the GridSearchCV class of the 
scikit-learn package to find the hyperparameters that provided the highest 
score.  We passed in all the possible default kernels of the SVC as well as 
powers of $10$ ranging from $10^{-4}$ to $10$ for the C hyperparameter as 
arguments to iterate over to find the parameter that provides the highest 
5-fold cross-validation
balanced accuracy score.

\subsubsection{Results}
\begin{figure}[ht]
	\centering
	\includegraphics[width=0.95\textwidth]{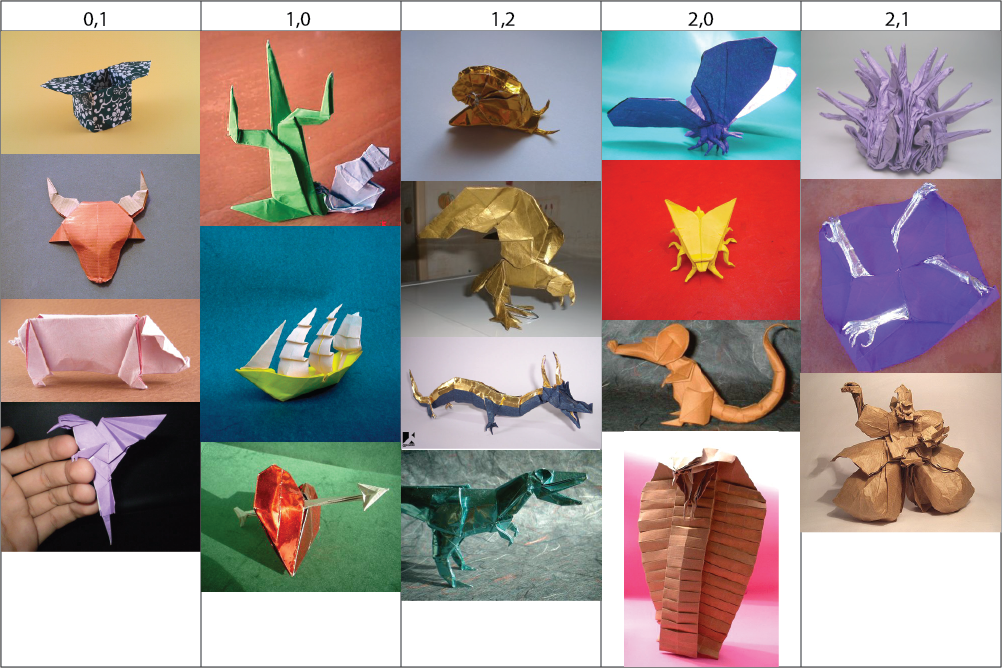}
	\caption{A collection of images that were wrongly classified.  The first 
	number in each column represents the labeled difficulty, while the second 
	number represents the predicted difficulty.}
	\label{f:wrong_preds}
\end{figure}

We found that an SVC with a polynomial kernel with $C=0.1$ provided the highest 
possible score, giving us a balanced accuracy of $0.7913\pm0.0219$. 
Figure~\ref{f:wrong_preds} shows several of the images that were classified 
incorrectly using these parameters for the SVC.

\subsubsection{Discussion}
Keeping in mind, that difficulty estimation is a challenging task
and that guessing would result in a balanced accuracy of $0.3333$,
our results show that a difficulty estimation with machine learning 
and our three difficulty levels is in general possible.

\section{Conclusion}
\label{s:conc}
In this paper, we introduced OrigamiSet1.0, which
provides images for the origami classification as well as
difficulty estimation.
Our empirical results show that our data can be successfully used for these two
machine learning challenges.

Future challenges are to estimate the number of folds that a model contains 
as well as to classify
what an origami model is supposed to represent.
Furthermore, the existing approaches can be improved in performance.
For example, randomizing background, lighting, and paper texture (if possible)
can be helpful augmentation techniques on the training data and also more
tailored classification algorithms can improve performance.
However, this requires an extension of the dataset in size as well as information content.
We hope this paper encourages the interested reader to contribute more data.

\section*{Acknowledgments}

This work was supported by a fellowship from the FITweltweit program 
of the German Academic Exchange Service (DAAD),
by the Undergraduate Research Apprenticeship Program (URAP) 
at University of California, Berkeley, and by grants from the U.S.\ National Science 
Foundation (1251276 and 1629990).
(Findings and conclusions are those of the authors, and do not necessarily
represent the views of the funders.)

\bibliographystyle{abbrv}
\bibliography{library,add_library}

\end{document}